\documentclass{article}

\usepackage[final]{nips_2017}
\usepackage{amsthm}
\usepackage{amssymb}
\usepackage{natbib}
\usepackage{mathtools}

\usepackage{algcompatible}
\usepackage{hyperref}
\usepackage[utf8]{inputenc} % allow utf-8 input
\usepackage[T1]{fontenc}    % use 8-bit T1 fonts
\usepackage{hyperref}       % hyperlinks
\usepackage{url}            % simple URL typesetting
\usepackage{booktabs}       % professional-quality tables
\usepackage{amsfonts}       % blackboard math symbols
\usepackage{nicefrac}       % compact symbols for 1/2, etc.
\usepackage{microtype}      % microtypography
\usepackage{wrapfig,lipsum,booktabs}
\usepackage{mathtools}
\usepackage{array}
\newcolumntype{P}[1]{>{\centering\arraybackslash}p{#1}}
\usepackage{times}
\usepackage{footnote}
\usepackage{tablefootnote}
\makesavenoteenv{tabular}
\makesavenoteenv{table}
\usepackage{epsfig}
\usepackage{graphicx}
\usepackage{sidecap}%
\usepackage{color}
\usepackage{amsmath, dsfont, bbm, bm, amsfonts}
\usepackage[normalem]{ulem}
\usepackage{soul}
\usepackage{amssymb}
\usepackage[T1]{fontenc}
\usepackage{hyperref} 
\usepackage{fixltx2e}
\usepackage{color}
\usepackage{algorithm}
\usepackage{algpseudocode}
\usepackage{caption}
\usepackage{subcaption}
\usepackage{subcaption,placeins,blindtext}
\usepackage{setspace}
\usepackage{epstopdf}
\usepackage[utf8]{inputenc}
\usepackage[english]{babel}

\floatstyle{plaintop}
\restylefloat{table}

\usepackage{caption}

\def\NoNumber#1{{\def\alglinenumber##1{}\State #1}\addtocounter{ALG@line}{-1}}

\def\balpha{\boldsymbol\alpha}

\def\bx{\mathbf{x}}

\def\bd{\mathbf{d}}
\def\bD{\mathbf{D}}

\def\bI{\mathbf{I}}

\def\bA{\mathbf{A}}

\def\bz{\mathbf{z}}

\def\bb{\mathbf{b}}
\def\bw{\mathbf{w}}
\def\bbR{{\mathbb R}}

\def\bgamma{\boldsymbol\gamma}

\sidecaptionvpos{figure}{c}

\makeatletter
\newcommand*{\rom}[1]{\expandafter\@slowromancap\romannumeral #1@}
\makeatother

 \newcommand{\norm}[1]{\lVert #1 \rVert}

\makeatother

\usepackage{array}
\makeatletter
\newcommand{\thickhline}{%
    \noalign {\ifnum 0=`}\fi \hrule height 1pt
    \futurelet \reserved@a \@xhline
}
\newcolumntype{"}{@{\hskip\tabcolsep\vrule width 1pt\hskip\tabcolsep}}
\makeatother
\usepackage{booktabs}

\title{Supervised Deep Sparse Coding Networks}

\author{
  Xiaoxia Sun$^1$, Nasser M. Nasrabadi$^2$, and Trac D. Tran$^1$\\
  The Johns Hopkins University$^1$\\
  West Virginia University$^2$\\ 
  \texttt{\{xxsun, trac\}@jhu.edu, nasser.nasrabadi@mail.wvu.edu} \\  
}

\begin{document}

\maketitle

\begin{abstract}
%Sparse coding represents a signal with a linear combination of  few dictionary atoms. Given an overcomplete dictionary, sparse coding is able to recover the high dimensional compact representations from  low dimensional feature space. In the case where dictionary is undercomplete, sparse coding along with the nonnegative constraint is able to encourage representations to form clusters, which is conceptually similar to the semi-nonnegative matrix factorization (semi-NMF). To the best of our knowledge, these two symbolic characters of sparse coding  have yet been simoultaneously exploited in  deep network. 

 In this paper,  we  describe the deep sparse coding network (SCN), a novel deep network  that encodes intermediate representations  with nonnegative sparse coding. The SCN is built upon a number of cascading {\it bottleneck modules}, where each module consists of two sparse coding layers with relatively wide and slim dictionaries that are specialized to produce high dimensional discriminative features and low dimensional representations for clustering, respectively.  During  training, both the dictionaries and regularization parameters are optimized with an end-to-end supervised learning algorithm based on multilevel optimization. Effectiveness of an SCN with seven bottleneck modules \footnote{Consisting 14 sparse coding layers.}  is verified on several popular benchmark datasets  \footnote{Our codes and models are available on \href{https://github.com/XiaoxiaSun/supervised-deep-sparse-coding-networks}{https://github.com/XiaoxiaSun/supervised-deep-sparse-coding-networks}. }.  Remarkably, with few parameters to learn, our SCN achieves $5.81\%$ and $19.93\%$ classification error rate on CIFAR-10 and CIFAR-100, respectively.
\end{abstract}

%\textcolor{red}{Main arguement of the paper is: 
%1. We propose bottleneck module with dimensionality expansion and reduction.
%2. Why use bottleneck module? Dimensionality expansion works like subspace detector. Dimensionality reduction is equivalent with clustering, give us the indicator of the subspace. 3. Bottleneck module is also used in neural network, what is the difference? Dimensionality reduction (in neural network) does not equal with clustering (in our case)!}

\section{Introduction}
Representing a  signal with a linear combination of few dictionary atoms, sparse coding has shown promising performance on a range of computer vision tasks including  image classification and object detection \cite{robustface,jwright,local_sparse_coding}. Even when given only a small amount of training samples, sparse coding models can become exceptionally resilient against severely corrupted or noisy data. However, when the noise in the data is  an expression of the natural variation of objects, such as those caused by changes in orientation, the linear representation of sparse coding is not able to catpure these nonlinear variations and contribute to the degradation of the performance. As such, sparse coding models exhibit disappointing performance on large datasets where variability is broad and anomalies are common.

 Conversely, deep neural networks thrive on large volume of data. Their success derives from an ability to distill the core essence of a subject from abundant diverse examples \cite{alexnet}, which has encouraged researchers to try and augment the learning capacity of traditionally shallow sparse coding methods by adding layers \cite{fast_sparse_coding,deep_sparse_coding,representation_learning_nc}.
 Multilayer sparse coding networks are expected to combine the best of both strategies. For instance, the imperative for sparse codes to adequately reconstruct an input signal ameliorates information degeneracy issues within deep  architectures. Furthermore, with parsimoneous representations, sparse coding networks can lead to more intuitive intepretation of the learned features.  From the view point of neural network, sparse coding enjoys a much stronger `explain away' effect \cite{bengio_representation_learning}  compared to the feedforward network and is closely related to  recurrent network through unfolding the sparse recovery algorithm \cite{fast_sparse_coding,iclr2017_understand_sparse_learning}.  To date, however, endeavors to marry the two techniques have not achieved significant improvements over their individual counterparts \cite{deep_sparse_coding,representation_learning_nc}.

In this paper, we present a novel deep sparse coding network  based on nonnegative sparse coding and multilevel optimization. { The goal of our work is to efficiently extend the conventional sparse coding to multilayer architectures in order to expand its learning capacity.}  We present a { bottleneck module}, the core building block of our SCN, which employs two specialized  sparse coding layers: one is  equiped with a relatively wide dictionary, while the other one has a slim dictionary  to significantly reduce the number of learnable parameters without incurring the penalty  on the performance of the network. The width of the sparse coding layer is defined as the number of dictionary atoms.    We also propose to optimize both the dictionaries and regularization parameters of SCN using an end-to-end supervised learning algorithm based on multilevel optimization. 

%  Moreoever, the proposed network is highly compatible with deep neural networks and the two types of networks are reciprocal. 

We demonstrate the effectiveness of the proposed SCN on four benchmark data sets, including CIFAR-$10$, CIFAR-$100$, STL-$10$ and MNIST.  The proposed network  exhibits competitive performance  using only a few learnable parameters. Remarkably, our $15$-layer \footnote{Including $14$ sparse coding layers and one linear classifier.} SCN achieve higher accuracy on CIFAR-10 and CIFAR-100 compared to a $110$-layer \cite{He_2016_CVPR} and $1001$-layer \cite{residual_1001} residual network, respectively. We emphasize here that in this paper  we do not aim at pursuing highest accuracy, but rather effectively extending sparse coding to multilayer architectures to achieve competitive performance.

\section{Related Works}
The proposed SCN is mostly related but substantially differs from the following approaches:

\textbf{Deep neural networks with wide shape.}  Wide networks \cite{wide_residual_networks,swapout,
ResNext} have been recently proposed to exploit the high dimensional latent features, demonstrating competitive performance compared to slim deep network \cite{He_2016_CVPR,residual_1001,SqueezeNet}  while using much fewer layers. The wide networks consist of a much larger number of learnable parameters compared to their slim counterparts, therefore requiring much stronger regularization to reduce overfitting. In this paper, we focus on regularizating the wide SCN via dimensionality reduction and clustering using the nonnegative sparse coding-based approach.

\textbf{Dimensionality reduction and clustering in deep neural networks.}  Bottleneck shaped neural network \cite{He_2016_CVPR,ResNext} applies dimensionality reduction in order  to reduce the overfitting of residual network. In contrast to neural network, dimensionality reduction with nonnegative sparse coding is equivalent to clustering \cite{seminmf} and therefore the low dimensional hidden features act as weighted cluster indicators which is discussed in Section \ref{sec:bottleneck_module}.   Alternate approach related to our work is the deep semi-nonnegative matrix factorization (semi-NMF) \cite{deep_semi_nmf} that trains a hierarchical network with the reconstruction loss. Our approach differs from the aforementioned works since we simoultaneously learn high dimensional discriminative representations and low dimensional clustered features in a single network architecture with end-to-end supervised learning.

\textbf{Multilayer sparse coding. } Most common approach for multilayer sparse coding is to train the reconstructive dictionaries of each layer in a greedily layer-wise fashion  \cite{representation_learning_nc,ICML2011Coates_485}, where the nonlinearity is usually enforced with a ReLU layer.   An alternative approach is to unfold and approximate the sparse coding process  with deep neural networks \cite{fast_sparse_coding,iclr2017_understand_sparse_learning}, in which the sparse coding parameters are trained end-to-end by minimizing reconstruction loss. Our approach differs in that we extend the conventional nonnegative sparse coding to a multilayer architecture without applying unfolding during training, therefore the model size would not increase with the number of iteration. Moreover, our approach allows a much broader choice among the off-the-shelf  sparse recovery algorithms without changing the network architecture. In addition, the proposed network also shares a high level motivation with the stacked autoencoder \cite{stacked_autoencoder} and CNN-based model with  auxiliary reconstruction loss \cite{SWWAE,yuting_zhang}, which trains the network in an unsupervised, semi-supervised or supervised fashion by manually balancing the discriminative and reconstruction loss. In contrast we employ conventional sparse coding instead of neural network to encode latent features and train the network supervisedly in an end-to-end fashion.

\textbf{Supervised dictionary learning.} Supervised dictionary learning strengthens the discriminative power of the sparse codes by exploiting the labeled samples.     Thorough study on task-driven dictionary learning algorithms for various applications is reported in \cite{MairalTDDL}. Applying fixed point differntiation and bilevel optimization, a supervised dictionary learning scheme for the shallow sparse coding model  is proposed in \cite{Yang,bilevelsc}.  In this paper, we generalize the single-layer supervised dictionary learning to multilayer network based on multilevel optimization.

\section{Extension of Sparse Coding to Deep Architectures}
\label{section:sparse_coding_networks}

\begin{figure*}[t]
	  \centering
	 % \begin{subfigure}[b]{0.60\textwidth}
	        \centering
	         \centerline{       \includegraphics[width=15cm]{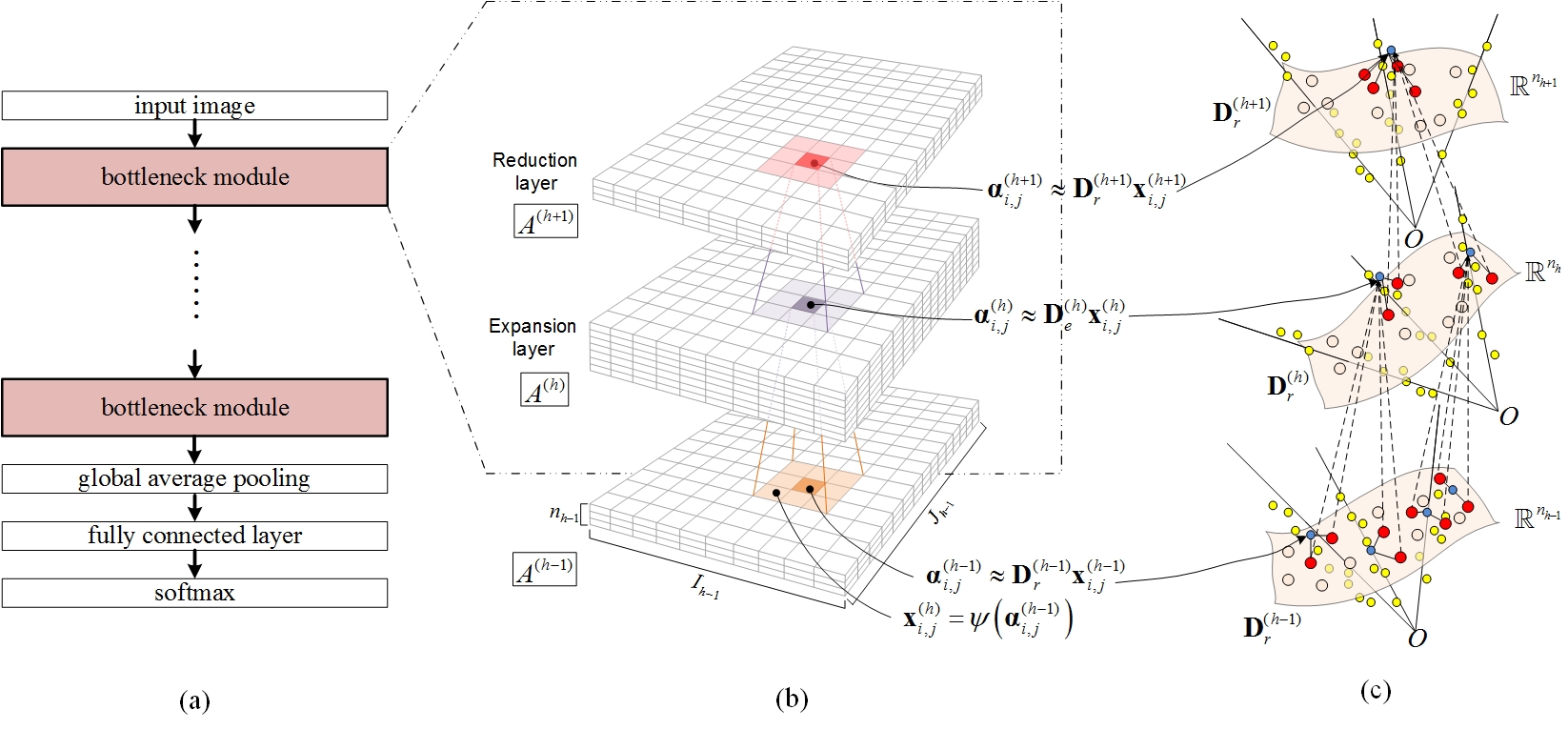}  }
	                %\caption{}
	                % \label{fig:module}
	 % \end{subfigure} \hspace{2em}
%	  \begin{subfigure}[b]{0.25\textwidth}
%	        \centering
%	         \centerline{       \includegraphics[width=2cm]{Figure/network_flow.png}  }
%	                \caption{}
%	                 \label{fig:flow_chart}
%	  \end{subfigure}  
	\captionsetup{justification=raggedright, singlelinecheck=false}
	
	\caption{\footnotesize Architecture of our multilayer sparse coding network:  (a) The proposed network is constructed by repeatedly stacking multiple bottleneck modules. The network does not contain any pooling operation and subsampling is conducted with a stride of $2$. (b) Bottleneck module consists of one expansion layer and one reduction layer, which is used to expand or reduce the dimensionality of the local features of the previous layer, respectively. (c) Intepretation of SCN. Red and hollow circles on the manifolds are the active and inactive atoms, respectively. Yellow circles are represents all the local features of a hidden layer and blue circles denotes the neighboring local features centered at $(i,j)$.  }
	\label{fig:networks_architecture}
	\end{figure*}

In this section, we introduce the SCN model by extending the conventional sparse coding to deep architectures. We start with the illustration of inference with nonnegative sparse coding which can be applied to every sparse coding layer in SCN. We then introduce the bottleneck module, where the sparse coding layers are specialized to encourage discriminative or clustered representations. Throughout this section, we assume that the dictionaries in SCN are given and we describe the dictionary learning algorithm  later in Section \ref{sec:dl}. The architecture of the SCN is illustrated in Fig. \ref{fig:networks_architecture}.

\subsection{Inference with Nonnegative Sparse Coding}
We now introduce a general formulation of sparse coding layer for SCN. Let the representation of the layer $h$ in SCN  be a $3$D-tensor ${\cal A}^{(h)} \in\bbR^{n_h\times I_{h} \times J_{h}}$, $h\in\{0, \dots, H\}$ and denote each local feature vector at $(i,j)$ of layer $h$ as $\balpha_{i,j}^{(h)} \triangleq {\cal A}^{(h)}_{:,i,j}\in\bbR^{n_h}$, where $n_h, I_{h}$ and $J_{h}$ are the number of channels, height and width of the layer representation.  For instance, $\balpha_{i,j}^{(0)}$ of a color image represents  a $3$-channel pixel of red, green and blue. In deeper layers where $h>0$, $\balpha_{i,j}^{(h)}$ represent a local sparse code. To  recover the local sparse code $\balpha_{i,j}^{(h)}$, we construct an intermediate local feature $\bx_{i,j}^{(h)}=\psi(\balpha_{i,j}^{(h-1)})\in\bbR^{m_h}$  by concatenating all the neighboring features centered at $(i, j)$ within a window of size $k_{h-1}\times k_{h-1}$ from  the previous layer $h-1$, For illustrative purpose, we assume the neighboring window is square. $\psi(\cdot)$ denotes the  concatenation operation and  $m_h = n_{h-1}k_{h-1}^2$.   We constrain the sparse codes to be nonnegative in order to introduce nonlinearity to the deep network.  Given a dictionary $\bD^{(h)}\in\bbR^{m_h\times n_{h}}$ of layer $h$,  the nonnegative sparse code is recovered by solving the following constrained elastic net problem:
\begin{equation}
\balpha^* =  \arg\min_{\balpha>\mathbf{0}}F(\balpha) \triangleq\arg\min_{\balpha>\mathbf{0}}\frac{1}{2}\norm{\bx - \bD\balpha}_2^2+\lambda_1\norm{\balpha}_1 + \frac{\lambda_2}{2}\norm{\balpha}_2^2,
\label{eq:nonnegative_sparse_coding}
\end{equation} 
where we have omitted the coordinate and layer indices  for simplicity.  $\norm{\balpha}_1 = \sum_{n=1}^N{|\alpha_n|}$ is the $\ell_1$-norm and $\lambda_1, \lambda_2>0$ are the regularization parameters.  Importance of the parameter $\lambda_2$ is to stabilize the training procedure \cite{MairalTDDL}.  {\it In this paper, we  directly solve  (\ref{eq:nonnegative_sparse_coding}) using conventional sparse recovery algorithm for inference instead of applying unfolding on sparse coding process with deep neural network} \cite{fast_sparse_coding,iclr2017_understand_sparse_learning}.  Number of sparse recovery algorithms such as LISTA \cite{fast_sparse_coding,iclr2017_understand_sparse_learning}, FISTA \cite{fista} and LARS \cite{Efron04leastangle} can efficiently solve problem \ref{eq:nonnegative_sparse_coding}. In this paper, we adopt FISTA mainly for the ease of coding in GPUs. The nonnegativity is enforced by using nonnegative soft-thresholding during the optimization.  For the purpose of clarity, sparse recovery algorithm for solving problem \ref{eq:nonnegative_sparse_coding} is shown in Appendix A.

\subsection{Bottleneck Module with Specialized Sparse Coding Layer}
\label{sec:bottleneck_module}
We now describe the {\it bottleneck module} which is the core building block of our SCN. Each bottleneck module consists of a cascade of two specialized sparse coding layers, which are referred to as {\it expansion layer} and {\it reduction layer}. The expansion layer is equiped with a relatively wide dictionary in order to reach a fine-grained partition of the input feature space, whereas the reduction layer has a relatively slim dictionary which focuses more on dimensionality reduction and clustering in order to extract more abstract representations.   We denote the dictionaries in expansion and reduction layer as $\bD_e\in\bbR^{m_e\times n_e}$ and $\bD_c\in\bbR^{m_r\times n_r}$, respectively, where $n_e\gg n_r$. We note that the order of expansion and reduction layer in a bottleneck module does not matter much in the multilayer environment. For illustrative purpose, we sequentially employ expansion layer and  reduction layer in a single bottleneck module.  We  illustrate the two specialized sparse coding layers and describe the motivations of proposing the bottleneck module in more details:

\textbf{Expansion layer focuses on partitioning feature space. } Nonnegative sparse coding  functions as a robust and stable partition of the input feature space \cite{Yang}, where the `resolution' of the partition depends on the dictionary width. With a relatively wide or even overcomplete dictionary, we are able to achieve a high resolution fine-grained partition of the feature space and therefore recover highly discriminative sparse codes. Behavior of sparse coding with a wide or even overcomplete dictionary  in single layer environment has been thoroughly exploited through number of studies \cite{jwright,MairalTDDL,Yang}.

%The high dimensional sparse codes recovered by expansion layer is usually highly uncorrelated in that similar input features can arrive at drastically different sub-coordinates. 

\textbf{Reduction layer focuses on clustering features. } 
Reduction layer  is designed to produce abstract compact sparse codes using a much narrower dictionary compared to that of the expansion layer. Output of the expansion layer has a high dimensionality and is intensely sparse,  which poses two issues in SCN. First, neighboring features from similar subspaces could be represented by highly distinctive sparse codes when the dictionary is overcomplete.  Second, high dimensional outputs  require the dictionaries in deeper layers to be excessively wide in order to maintain the redundancy and overcompleteness, which makes the network computationally infeasible. In this paper, inspired by the SqueezeNet \cite{SqueezeNet}, we propose to compress the high dimensional latent sparse codes using a reduction layer, which can be interpreted as conducting clustering on the high dimensional sparse codes.

 Nonnegative sparse coding with slim dictionary functions as clustering, which can be illustrated based on  semi-NMF \cite{seminmf}. Several inspirational works \cite{iclr2017_understand_sparse_learning,online_mairal,nonnegative_sc} illustrate the relations between sparse coding, dictionary learning and matrix factorization: In a reduction layer, when the given dictionary is slim, the nonnegative sparse coding is equivalent with sparsity-regularized semi-NMF algorithm, which is strongly related to the K-means clustering.  Hence, the slim dictionary atoms in reduction layer can be interpreted as the cluster centroids of the high dimensional inputs, whereas the corresponding low dimensional nonnegative sparse code is the weighted cluster indicator.

\subsection{Interpreting Sparse Coding Network as Deep Subspace Learning}
%In sparse coding-based model, it is natural to assume that the local feature descriptors in layer $h$ lie in a small subset of subspaces in $\bbR^{n_h}$ \cite{robustface}. 

For illustrative purposes, we consider the simplified case where all  the local features of layer $h$ lie on a union of disjoint subspaces, i.e., every pair of these subspaces only intersect at origin. As is shown in Fig. \ref{fig:networks_architecture} (c), each atom of the learned dictionary is the cluster center of a large number of local features in $\bbR^{n_{h}}$ and every nonnegative sparse code $\balpha^{(h)}_{i,j}$ in layer $h$ describes how strong it is connected to a certain cluster center.   We note that in the case of supervised learning, the distance between each local feature and their related cluster centers, i.e., dictionary atoms,  are not only measured by the reconstructive loss but also described by the discriminative loss as shown in (\ref{eq:supervised_dl}). 

In the case of SCN, large number of subspaces in $\bbR^{n_h}$ are related to each other through the local sparse code $\balpha^{(h)}_{i,j}$, which  itself lies on another subspace in $\bbR^{n_{h+1}}$ of the deeper layer $h+1$. Similarly, as the network goes deeper, each point in $\bbR^{n_{k}}$ of layer $k$ relates to a more complex union of subspaces in  $\bbR^{n_{j}}$ of the shallower layer $j$, where $k\gg j$, i.e., local sparse codes in deeper layers are more expressive compared to those from shallower layers.  Driven by the discriminative loss function, the local features  of two different classes are gradually mapped to different subspaces of each layer and eventually become linearly separable with respect to the hyperplane defined by the classifier.

%  When exploiting neighboring feature, local features that come from more than one subspace are mapped to single point, which belongs to a single subspace in the deeper layer. Note that neighboring local features cannot belong to the same linear subspace in layer $h$ since translation is nonlinear transformation. Hence, numerous points belongs to numerous subspaces in $\bbR^{n_{h}}$ in layer $h$ relates with a single point of subspaces in layer $h$. If the sparse code is perfectly recovered, then the network is a non-inject function.  

\subsection{Sparse Coding Network Structure}
  
Our SCN is designed to stack multiple bottleneck modules in order to perform dimensionality expansion and reduction repeatedly.  Batch normalization layer \cite{batch_normalization} is added after each sparse coding layer in order to obtain a faster convergence. The last bottleneck module lies on top of a global average pooling layer, which is followed by a fully connected layer which functions as the linear classifier.
%
%More specifically, $\bx^{(h)}={\cal F}(\bx^{(h-1)})$ for a local feature. When training the dictionary, we would like the dictionary to better represent the signal at each space ${\cal X}^{(h)}$, which facilitates a better generalization,  while the upper level goal is to achieve a better class decision in the last layer. Moreover, we also needs to optimize the regularization parameters in each layer since choose with cross-validation in the large parameters space of deep architecture is almost intimidating.

\section{End-to-end Supervised Learning for Sparse Coding Networks }
\label{sec:dl}
In previous section, we described the architecture of SCN with the bottleneck module. In this section,  we develop the end-to-end supervised learning algorithm for training both the dictionaries and regularization parameters in the deep SCN based on multilevel optimization. 

\subsection{Problem Formulation with Multilevel Optimization}
Without loss of generality, we consider a prediction task for binary class given a set of training pairs $\{{{\cal A}^{(0)}_s}, y_s\}_{s=1}^S$, where $y_s\in\{0,1\}$ is the label for the image sample ${\cal A}^{(0)}_s$. Given an SCN with $H$ sparse coding layers, our goal is to fit the network prediction to the label through minimizing a smooth and convex  loss function $L: \bbR\times\bbR\rightarrow \bbR$ with respect to the network parameters, including dictionaries, regularization parameters and the linear classifier. Suppose the network maps the input image ${\cal A}^{(0)}_s$ to the corresponding label $y_s$ with a nonlinear function $f: \bbR^{n_0}\times\bbR^{I_0}\times\bbR^{J_0}\rightarrow \bbR$,  the optimization procedure of  SCN is formulated as an empirical risk minimization problem based on multilevel optimization:
\begin{align}
\scriptsize
&\min_{\theta}  \frac{1}{S}\sum_{s=1}^S L(y_s, f({\cal A}^{(h)}_s, \bw)) + \frac{\mu}{2} R(\theta), \nonumber \\
&s.t.\quad \balpha_s^{{(H)}^*} = \arg\min_{\balpha_s^{(H)}\geq \mathbf{0}}F(\bD^{(H)}, \lambda^{(H)}, \bx^{(H)_s}, \balpha_s^{(H)}), \nonumber \\
&\quad \quad \quad \vdots \nonumber \\
& \quad s.t. \quad \balpha_s^{{(1)}^*} = \arg\min_{\balpha^{(1)}\geq \mathbf{0}}F(\bD^{(1)}, \lambda^{(1)}, \bx^{(1)}_s, \balpha^{(1)}_s), \nonumber \\
& \quad\quad\quad s.t. \quad   \lambda^{(h)}> 0, \;\bx_s^{(h)} = \psi(\balpha_s^{(h-1)^*}), \quad\forall h=1,\dots, H,
\label{eq:supervised_dl}
\end{align}
where  $\theta = \{\bD^{(h)}, \lambda^{(h)}\}_{h=1}^H$ is the learnable parameter set including both dictionaris and regularization parameters. $\bw\in\bbR^{n_H}$ is the linear classifier. In this paper, we adaptively optimize the regularization parameters at each layer, which has a similar effect as training the bias in deep neural networks \cite{iclr2017_understand_sparse_learning}.  The motivation for training regularization parameters is that cross-validation becomes formidable as the network becomes deeper.   

To prevent the $\ell_2$-norm of dictionary to be arbitrarily large and recovering trivial sparse codes, we introduce regularizer $R(\bD) \triangleq \norm{\bD}_F^2$, or usually referred to as weight decay in deep neural network, on the dictionary   to reduce the overfitting. We note that constraining every dictionary atom with $\norm{\bd_j}_2\leq c$, where $c>0$ is a chosen constant, is the most common choice for regularizing dictionary atoms in a single layer model. However, during experiment, we found that such constraint is too stringent for the network to converge due to the projection on descent gradient.

%the most common way for regularize the dictionary in the case of single-layer model is to . However, in the case of deep SCN, such constraint is too stringent for the network to converge: First, choosing the desired constant $c$ is not trivial in deep networks \cite{weight_init}. Second, projecting the gradient onto the feasible set introduces noise in the gradient, which accumulates as the network becomes deeper and deter the convengence. Hence,   In this paper, we exploit two different type of regularization functions on SCN:

%\textbf{Regularize the dictionary with $\ell_F$-norm. } Regularize the dictionary with Frobenious norm is the most common choice in deep networks, i.e., $R(\bD) \triangleq \norm{\bD}_F^2$. It is a strong regularization that effeciently suppresses the $\ell_2$-norm of every dictionary atom. 

\subsection{Updating Dictionary and Regularization Parameter}
\label{sec:dictionary_update}
Every sparse code $\balpha$ is parameterized by the dictionary and regularization parameters, it is therefore natural to solve the multilevel optimization problem (\ref{eq:supervised_dl}) with  gradient descent method based on error backpropagation \cite{bilevel_survey}. The derivation of the updating rules is based on the fixed point differentiation \cite{MairalTDDL,Yang,bilevelsc}. Due to the page limitation, we state the  first order optimality condition of the nonnegative elastic net, which is the core building block of the derivation, and we leave the rest parts in Appendix B.

\textbf{Lemma 1 (Optimality conditions of nonnegative elastic net.)} {\it \quad The optimal sparse code $\balpha^*$  of (\ref{eq:nonnegative_sparse_coding})   solves the following system:}
\begin{align}
    &\bd_j^\top(\bD\balpha^* - \bx) + \lambda_2\alpha^*_j  = -\lambda_1, \;\text{if} \;  \alpha^*_j> 0  \label{eq:stationary_cond1}\\
   &\bd_j^\top(\bD\balpha^* - \bx) + \lambda_2\alpha^*_j \geq -\lambda_1 , \; \text{otherwise}.
\label{eq:stationary_cond2}
\end{align}
{\it The nonnegative part of the sparse code $\balpha^*$ can be described as  $\balpha_\Lambda^* = (\bD_\Lambda^\top\bD_\Lambda + \lambda_2\bI)^{-1}(\bD_\Lambda^\top\bx - \lambda_1\mathbf{1}_\Lambda)$, where $\mathbf{1}_\Lambda\in\bbR^{|\Lambda|}$ is an all one vector, $\Lambda$ is the active set of $\balpha^*$ and $|\Lambda|$ is the cardinality of the active set $\Lambda$.}
\begin{proof}
Let $\partial\norm{\balpha}_1$ be the subgradient of $\norm{\balpha}_1$. Since  $\balpha^*$ is a the optimum of (\ref{eq:nonnegative_sparse_coding}),  $\forall j\in[N], \exists\mathbf{\bz}\in\partial \norm{\balpha}_1$, such that $\balpha^*$ solves the nonlinear Karush–Kuhn–Tucker (KKT) system  $(\bd_j^\top(\bD\balpha^* - \bx)  + \lambda_2\alpha_j^* + \lambda_1 z_j)\cdot \alpha_j=0$ and $z_j$ is the $j^{\text{th}}$ element of $\bz$. Followed by the classical result of Elastic Net \cite{MairalTDDL} and the complementary slackness of KKT condition,  when  $\alpha_j=0$, we have $z_j\leq 1$ and reach  (\ref{eq:stationary_cond1}). Representation of $\balpha_\Lambda^*$ can be achieved by applying algebraic simplification on (\ref{eq:stationary_cond1}) for all atoms $j$.  When  $\alpha_j>0$, we have $z_j=1$ and reach (\ref{eq:stationary_cond2}).   
\end{proof}

%Differentiation of the sparse codes with respect to the dictionary has been thoroughly studied in a series of works \cite{MairalTDDL,bilevelsc,Yang}. Inspired by previous works, the backpropagation rule  for updating dictionary and regularization parameters of SCN is derived from Lemma 1:

%Given above Eq. \ref{eq:nonnegative_sparse_coding}, one can approximate the input signal $\balpha^{(0)}$ with the latent sparse codes from any layer:
%\begin{equation}
%\balpha^{(0)} = \prod_{h=1}^k \bD^{(h)}\balpha^{(k)} + \be^{(k)},\quad \forall k\in[H].
%\end{equation}

%
%\subsection{Time complexity}
%The time complexity for inference with sparse recovery using FISTA is $\mathcal{O}(N^2)$
%

% In the case of shallow sparse coding models, active atoms are usually defined as $\{x_i:|x_i|>\epsilon, \forall i\in[N]\}$, where $\epsilon$ is a small constant value to avoid numerical instability and $x_i$ is the $i$th element of the sparse code $\bx$. In multilayer sparse coding networks, a fixed threshold $\epsilon$ does not function well since the magnitude of the sparse codes changes drastically from one layer to another due to the lack of normalization on dictionaries atoms. To compensate for this effect, we set the threshold as
%\begin{align}
%\epsilon = \epsilon_0 \frac{\norm{y}_2}{\norm{A_i}_2},
%\label{eq:safe_guard}
%\end{align}
%where $i$ is the index of a dictionary atom with largest $\ell_2$ norm and $\epsilon_0$ is the threshold when both dictionary atoms and input signal are $\ell_2$-normalized. We set $\epsilon_0=10^{-3}$ throughout our paper.

\section{Experiments}
\label{section:experiments}

We conduct extensive experiments on CIFAR-10, CIFAR-100, STL-10 and MNIST. We demonstrate that the proposed  SCN exhibits competitive performance while using much smaller number of parameters and layers compared to numerous deep neural network approaches. Notably, a 14-layer SCN model  exceeds the performance of  a $164$-layer and $1001$-layer deep residual network, respectively. The proposed SCN is implemented  using Matlab with C++ and GPU backend based on the framework of MatConvNet \cite{vedaldi15matconvnet}. Experiments are conducted on a server with $4$ Nvidia Tesla P40 GPUs. 

\textbf{Model. }
The architecture of the network is similar to the ResNet \cite{He_2016_CVPR}. The SCN is configured with  $7$ bottleneck module which  includes $14$ sparse coding layers that are divided into $3$ sections, i.e., $(16, 16K)\times 3-(32, 32K)\times 2 - (64, 64K)\times 2$, where each $(M, MK)\times P$ denotes a bottleneck module that is repeatedly stacked for $P$ times,  the output dimensions of reduction and expansion layers are $M$ and $MK$.  For CIFAR-10 and CIFAR-100, we exploit the performance of the network with different width, i.e., $K\in\{1, 2, 4\}$. For MNIST and STL-10, we set $K=4$. We denote an SCN with width of $MK$ as SCN-K. The window size $k_h$ at each sparse coding layer has a size of $3\times 3$. We use the same network configurations for CIFAR-10, CIFAR-100 and STL-10, and for MNIST we set the number of filters at the first layer to be $8$ due to the simplicity of the dataset. Following the architecture of ResNet, we apply spatial subsampling with a factor of $2$ at the last two bottleneck modules. The last sparse coding layer is  followed by one global spatial average pooling layer \cite{nin} and one fully connected layer which is the linear classifier.  Batch normalization is added after each sparse coding layer  to facilitate the convergence.

\textbf{Training. } At the training stage, we apply data augmentation and preprocessing for all datasets except for MNIST with random horizontal flipping and random translation. The image is translated up to $4$ pixels in each direction for CIFAR-10 and CIFAR-100, and up to $12$ pixels for STL-10. Images in the same batch share the same augmentation parameters.   Both training and testing images are preprocessed with per-pixel-mean subtraction, which is a common procedure for preprocessing these datasets \cite{He_2016_CVPR,ICML2011Coates_485,nin,dsn}. We use a minibatch size of $128$ for MNIST, CIFAR-10, and CIFAR-100. For STL-10, we use a batch size of $16$ in order to have more iterations per epoch on the small training set. For all dataset, the initial learning rate is set to $0.1$ and SCN is trained with a total of $200$ epochs. For CIFAR-10, CIFAR-100 and STL-10, we follow a similar learning rate schedule with \cite{residual_1001}, where the learning rate decreases twice at $80$ and $160$ epochs by a factor of $10$. For MNIST, the network is trained with $25$ epochs, where the learning rate decreases at $10$ and $20$ epochs by a factor of $10$. The weight decay is set to $0.0005$ for all the dataset with cross-validation. 

%We only tune the initial learning rate by cross-validation.  We use the first $45,000$ samples for training and the remaining $5,000$ samples for testing. The weight decay is set to $0.0001$ and the initial learning rate is set to $0.01$. The learning rate is decreased by a factor of $10$ after $80$ epochs. We run a total number of $200$ epochs which takes about $35$ hours on our server.   Since we only tune the initial learning rate, we do not guarantee that our multilayer sparse coding network or the baseline CNN can reach its best performance.  

\textbf{Baseline comparison methods. } We compare our proposed SCN with numerous multilayer sparse coding-based approaches, including multilayer sparsity regularized coding (OMP) \cite{ICML2011Coates_485} and nonnegative multilayer sparse coding (NOMP) \cite{representation_learning_nc}. We also compare with supervised convolutional kernel networks (SCKN) \cite{sckn} and scattering network (ScatNet) \cite{scatnet}. For deep neural network baseline, we mainly compare with residual network (ResNet) \cite{He_2016_CVPR,residual_1001}, wide residual network (WRN) \cite{wide_residual_networks} and swapout networks (SwapOut) \cite{swapout}.

\textbf{Computation time. } For CIFAR-10 and CIFAR-100, training SCN-4 model with $200$ epochs takes about $26$ hours and inference of all the $10,000$ testing images takes about $9$ seconds. Training and inference with SCN-4 on STL-10 dataset takes about $21$ hours and $65$ seconds, respectively. For MNIST, training and testing takes about $3$ hours and $7$ seconds, respectively.

%	  \begin{figure}[t]
%	  
%	  \centering
%	         \centerline{       \includegraphics[width=8.7cm]{Figure/feature_map.jpg}  }
%	\captionsetup{justification=raggedright, singlelinecheck=false}
%	
%	\caption{\footnotesize Visualization of feature map: From left to right: Original image; feature maps of our sparse coding network - feature maps contain mostly background are labeled with yellow rectangles; and feature map of the baseline CNN. }
%	\label{fig:feature_map}
%	\end{figure}

\subsection{CIFAR-10 and CIFAR-100}

	  \begin{figure*}[ht!]
	  \centering
	  \begin{subfigure}[b]{.4\textwidth}
	        \centering
	         \centerline{       \includegraphics[width=6cm]{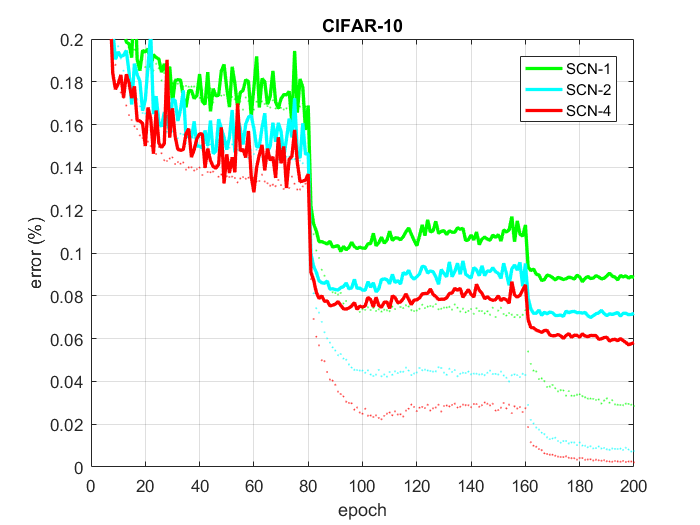}  }
	                \caption{}
	                 \label{fig:subsample}
	  \end{subfigure} \hspace{2em}
	  \begin{subfigure}[b]{.4\textwidth}
	        \centering
	         \centerline{       \includegraphics[width=6cm]{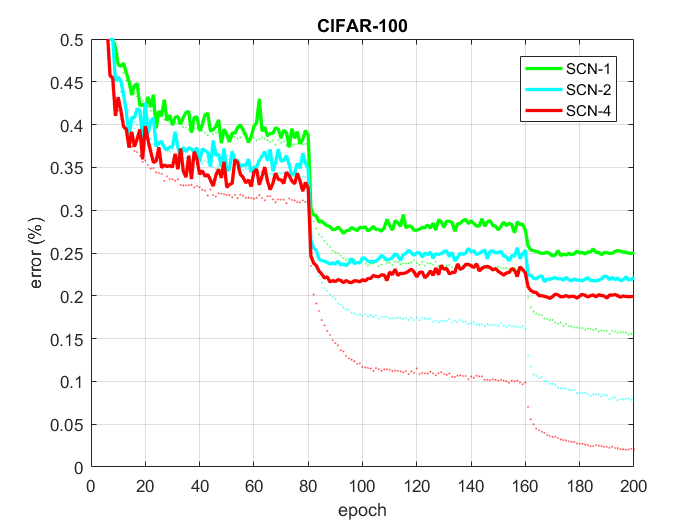}  }
	                \caption{}
	                 \label{fig:translation}
	  \end{subfigure} \\
	\captionsetup{justification=raggedright, singlelinecheck=false}
	
	\caption{Learning curve of SCN on CIFAR-10 and CIFAR-100. Dotted and solid lines denote the learning curves of training and testing stage, respectively.  }
	\label{fig:learning_curve}
	\end{figure*}

Our most extensive experiment is conducted on the CIFAR-10 and CIFAR-100 datasets, which consists of $60,000$ color images that are evenly splitted into $10$ classes.  The database is split into $50,000$ training samples and $10,000$ test samples. Each class has $5,000$ training images and $1,000$ testing images with size $32\times 32$.  CIFAR-100 has exactly the same set of images as CIFAR-10 but are split into $10$ times more classes, therefore each class has much fewer training samples compared with CIFAR-10, making it a more challenging dataset for the task of classification.

\begin{table}[ht]
\captionsetup{font=scriptsize}
\begin{center}
\scriptsize
\begin{tabular}{ >{\centering\arraybackslash}P{1.5in}|P{0.5in}|P{0.5in}|P{0.8in} |P{0.8in} }
\hline
{Method} & {\# Params} & {\# Layers}  & CIFAR-10 & CIFAR-100 \\
\hline
SCKN \cite{sckn} &$10.50$M &$10$ & $10.20$  & -\\
OMP \cite{ICML2011Coates_485} &$0.70$M &$2$ &$18.50$ & - \\
PCANet \cite{pcanet} &$0.28$B &$3$  &$21.33$ & -\\
NOMP \cite{representation_learning_nc} &$1.09$B &$4$ &$18.60$ & $39.92$ \\
NiN \cite{nin} &- &-  & $8.81$  & $35.68$ \\
DSN \cite{dsn} &$1.34$M &$7$ & $7.97$ &  $36.54$ \\
\hline
WRN \cite{wide_residual_networks}  & ${36.5}$M  &${28}$  & ${4.00}$ & ${19.25}$ \\
ResNet-110 \cite{He_2016_CVPR}  &$0.85$M &$110$ & $6.41$ & $27.22$ \\
ResNet-1001 v2 \cite{residual_1001}  & $10.2$M  &$1001$  & $4.92$ &  $27.21$ \\
\textbf{ResNext-29} \cite{ResNext}  & $\mathbf{68.10}$\textbf{M}  &$\mathbf{29}$  & $\mathbf{3.58}$ &  $\mathbf{17.31}$ \\
SwapOut-20 \cite{swapout}  &$1.10$M  & $20$ & $5.68$ & $25.86$ \\
SwapOut-32 \cite{swapout}  &$7.43$M  & $32$ & $4.76$ & $22.72$\\
\hline
%\textbf{Hybrid Network} &$\mathbf{0.47}$\textbf{M} &$\mathbf{14}$ & $\mathbf{6.55}$\\
{SCN-1} &${0.17}${M} &${15}$ & $8.86$  &$25.08$ \\
{SCN-2} &${0.35}${M} &${15}$ & $7.18$   & $22.17$\\
\textbf{SCN-4} &$\mathbf{0.69}$\textbf{M} &$\mathbf{15}$ & $\mathbf{5.81}$ & $\mathbf{19.93}$\\
\hline
\end{tabular}
\end{center}
\caption{Classification Error ($\%$) on CIFAR-10 and CIFAR-100.}
\label{table:cifar10}
\end{table}

\textbf{SCN outperforms previous deep sparse coding models. } As shown in Table \ref{table:cifar10},  for CIFAR-10, previous sparse coding-based models of OMP \cite{ICML2011Coates_485} and NOMP \cite{representation_learning_nc} networks reported a classification error of $18.50\%$ and $18.60\%$, respectively. With the aid of bottleneck module and end-to-end supervised learning algorithm, the proposed SCN demonstrates an improvement of $13\%$ on accuracy while only using $0.69$M parameters.

\textbf{Wider dictionary in expansion layer significantly improves performance. } 
Table \ref{table:cifar10} and Fig. \ref{fig:learning_curve} show that the classification performance of SCN increases with the width of the dictionary of the expansion layer, gaining $3\%$ and $6\%$ on CIFAR-10 and CIFAR-100, respectively. In the case when $K=4$, our $15$-layer SCN exhibits competitive performance compared to $20$-layer SwapOut network on CIFAR-10 while using twice fewer parameters.

\textbf{SCN with bottleneck module uses parameters efficiently. } From Table \ref{table:cifar10}, we can see that the proposed SCN uses fewest learnable parameters compared to {\it all} baseline models and contain fewest number of layers compared to all deep neural network-based baselines. Compared to the state-of-the-art approach of ResNext, our model uses almost $100\times$ fewer parameters and almost one half of layers while still reaching a  competitive performance. Moreover, the SCN-4 outperforms other approaches with similar model size such as ResNet-1001 on CIFAR-100.

\textbf{SCN exhibits strongly competitive performance compared to baselines models.} The proposed SCN achieves classification error of $5.81\%$ and $19.93\%$ on CIFAR-10 and CIFAR-100, respectively, which is shown in Table \ref{table:cifar10}. Consider the small size of our model, the performance of SCN is rather strong and competitive. In addition, our model has not yet exploited numerous useful tools in deep neural network such as dropout, shortcut connection and swapout, we strongly believe that the performance of SCN can be further improved.

\subsection{STL-10}

\begin{wraptable}{tr}{6.0cm}
\captionsetup{font=scriptsize}
\begin{center}
\scriptsize
\begin{tabular}{ >{\centering\arraybackslash}P{0.75in}|P{0.3in}|P{0.25in}|P{0.35in} }
\hline
{Method} & {\#Params} & {\#Layers}  & Accuracy \\
\hline
SWWAE \cite{SWWAE} &$10.50$M &$10$ & $74.33$ \\
{Deep-TEN} &${25.60}${M} &${50}$ & $76.29$   \\
\textbf{SCN-4} &$\mathbf{0.69}$\textbf{M} &$\mathbf{15}$ & $\mathbf{83.11}$ \\
\hline
\end{tabular}
\end{center}
\caption{Classification Accuracy ($\%$) on STL-10.}
\label{table:stl10}
\end{wraptable}

The dataset STL-$10$ is originally designed for unsupervised learning, which contains a total number of $5,000$ labeled training images and $8,000$ testing images with size of $96\times 96$. For this dataset, we follow the evaluation protocol used in \cite{zhang2016deep}.  The training samples in STL-10 is highly limited and SCN is supposed to generate more competitive performance due to the regularization from  the bottleneck modules. We directly apply the $14$ sparse coding layer model on STL-10 and replace the $8\times 8$ average pooling with $24\times 24$. We compare our network with the baseline of DeepTEN and previous state-of-the-art approach \cite{SWWAE}. From table \ref{table:stl10}, we can see SCN with bottleneck module (SCN-4) outperforms Deep-TEN under fair comparison by a large margin of $7\%$. SCN also exceeds previous state-of-the-art performance \cite{SWWAE} by almost $9\%$.

%\begin{table}[t]
%\begin{center}
%\begin{tabular}{>{\arraybackslash}m{1.7in} >{\centering\arraybackslash}m{1in} }
%\textbf{Method} & \textbf{Accuracy (\bm{\%})} \\
%\toprule
%CKN-PM \cite{CKN}  &$60.25$ \\
%Lin \emph{et. al.} - $1$ Layer \cite{representation_learning_nc} &$59.00$\\
%Lin \emph{et. al.} - $2$ Layers \cite{representation_learning_nc} &${60.40}$\\
%Sohn \emph{et. al.} \cite{ICML2012Sohn_659} & $58.70$\\
%Proposed - $14$ Layers  &  $\mathbf{83.11}$\\
%\hline
%\end{tabular}
%\end{center}
%\caption{STL-10 Classification Accuracy. }
%\label{table:stl10}
%\end{table}

\subsection{MNIST}

\begin{wraptable}{tr}{6.0cm}
\captionsetup{font=scriptsize}
\begin{center}
\scriptsize
\begin{tabular}{ >{\centering\arraybackslash}P{0.6in}|P{0.3in}|P{0.25in}|P{0.35in} }
{Method} & {\#Params} & {\#Layers}  & Accuracy \\
\hline
CKN \cite{CKN}  &  & $2$ &  $0.39$ \\
ScatNet \cite{scatnet} &- &$3$ & $0.43$ \\
PCANet \cite{pcanet} &- &$3$ & $0.62$ \\
S-SC \cite{Yang} &- &$1$ & $0.84$ \\
TDDL \cite{MairalTDDL} &- &$1$ & $0.54$ \\
%Highway \cite{highway} &$0.15$M & $20$  &  $0.45$ \\
\textbf{SCN-4} &$\mathbf{0.69}$\textbf{M} &$\mathbf{15}$ & $\mathbf{0.36}$\\
\hline
\end{tabular}
\end{center}
\caption{Classification Error ($\%$) on MNIST. }
\label{table:mnist}
\end{wraptable}

The MNIST dataset consists of $70,000$ images of digits, of which $60,000$ are the training set and the remaining $10,000$ are the test set. Each digit is centered and normalized to a $28\times28$ field.  The classification error on this dataset is reported in Table \ref{table:mnist}. { The purpose of evaluating this relatively simple dataset is to compare with previous sparse coding-based models, as most of which are evaluated on MNIST.}   With a deep architecture, the SCN achieves a classification error of $0.36\%$, which outperforms all previous sparse coding-based models.

%\subsection{Empircal Runtime Analysis}

%
%
\section{Conclusion and Discussion}
In this paper, we have developed a novel multilayer sparse coding network based on nonnegative sparse coding and multilevel optimization.  We propose applying bottleneck module  to dramatically reduce the overfitting and computational costs of SCN. Moreover, we also show that our SCN is compatible with other powerful deep learning tools such as batch normalization.  We have demonstrated that our network produces results competitive with deep neural networks but uses significantly fewer parameters and layers. 
{
\footnotesize
\bibliographystyle{unsrt}
\bibliography{refs}
}

\newpage

\section*{A \quad Solving Constrained Elastic Net using FISTA}

For the purpose of clarification, we describe the nonnegative FISTA in Algorithm \ref{algorithm:fista}, which is used for inference during training and testing. We denote $(\bA)_+$ as the element-wise nonnegative thresholding on $\bA$.

\begin{algorithm}[h]
%\small
\caption{FISTA for nonnegative Elastic Net}. 
\begin{algorithmic}[1]
%\caption{Task driven dictionary learning using group sparse recovery}
%\fontsize{7}{7}
% \caption{Task driven dictionary learning with joint sparsity prior}

 %\SetAlgoLined
% input: Initial $\bD$ and $\bW$, training data set $(\by, \bx)$, initial step size $\rho_0$.

\Require Dictionary $\bD\in\bbR^{M\times N}$, $\kappa$ is the largest eigenvalue of $(\bD^\top\bD+\lambda_2\bI)$, precompute $\bA=\bI - \frac{1}{\kappa}(\bD^\top\bD+\lambda_2\bI)$, $\bb=\frac{1}{\kappa}(\bD^\top\bx - \lambda_1)$, iterator $t=0$, $\balpha_t=\mathbf{0}\in\bbR^N$, $\bgamma_t=\mathbf{0}\in\bbR^N$, $s_0=1$.
 \While{stopping criterion not satisfied} 
 \State $\balpha_{t+1}\leftarrow (\bA\bgamma_t+\bb)_{+}$.
 \State $s_{t+1}\leftarrow(1+(1+4s_t^2))/2$.% t_new = 0.5*(1 + sqrt(1 + 4*t_old^2));
 \State $\bgamma_{t+1}\leftarrow \balpha_{t+1}+(s_{t}-1)(\balpha_{t+1}-\bx_t)/s_{t+1}$.  
 \State $t\leftarrow t+1$. %y_new = x_new + (t_old - 1)/t_new * (x_new - x_old); 4*t_old^2));
\EndWhile \\
\Return Nonnegative sparse code $\balpha_t$.

\end{algorithmic}
\label{algorithm:fista}
\end{algorithm}

\section*{B \quad Dictionary and parameter update}

We now derive the backpropagation rule for solving problem \ref{eq:supervised_dl}. In this paper, we derive the updating rule for the case of holistic sparse coding since extension to the convolutional local sparse coding is trivial. Optimizing dictionaries and regulerization parameters with stochastic gradient descent is shown in Algorithm \ref{algorithm:dl}.  We start by differentiating the empirical loss function with respect to every element of the dictionaries and regularization parameters:
\begin{align}
\frac{\partial L}{\partial d^{(h)}_{jk}} &=\frac{\partial L}{\partial \balpha^{(H)}}\cdot\left(\prod_{i=H}^{h+1}\frac{\partial \balpha^{{(i)}}}{\partial \balpha^{{(i-1)}}}\right)\cdot\frac{\partial \balpha^{{(h)}}}{\partial d_{jk}^{(h)}}, \label{eq:bp_0} \\
\frac{\partial L}{\partial \lambda_1^{(h)}} &= \frac{\partial L}{\partial \balpha^{(H)}}\cdot\left(\prod_{i=H}^{h+1}\frac{\partial \balpha^{{(i)}}}{\partial \balpha^{{(i-1)}}}\right)\cdot\frac{\partial \balpha^{{(h)}}}{\partial \lambda_1^{(h)}} , \quad \text{s.t.} \quad \lambda_1^{(h)} > 0,
\label{eq:bp_1}
\end{align} 
where $d_{jk}^{(h)}$ is the $(j,k)$-element of the dictionary $\bD^{(h)}$.  To solve for (\ref{eq:bp_0}) and (\ref{eq:bp_1}), we need to derive $\partial \balpha^{{(h)}}/\partial \balpha^{{(h-1)}}$, $\partial \balpha^{{(h)}}/\partial d_{jk}^{{(h)}}$ and $\partial \balpha^{{(h)}}/\partial \lambda_1^{{(h)}}$. We employ fixed point differentiation for deriving the required derivatives, which is based on the previous works of dictionary learning for one-layer sparse coding model  \cite{MairalTDDL,Yang}. Let $\balpha\in\bbR^N$ \footnote{We have omitted the superscript `*'  for simplicity.} be the optimal point of Lasso problem, it then satisfies the optimality condition based on  (\ref{eq:stationary_cond1}) and for all $\balpha_{\Lambda}>0$:
\begin{align}
(\bD^\top_\Lambda\bD_\Lambda+\lambda_2\bI_{|\Lambda|})\balpha_{\Lambda} - \bD_\Lambda^\top\bx + \lambda_1 \mathbf{1}_{|\Lambda|} = \mathbf{0},
\label{eq:explicit}
\end{align}%
where we have omitted the layer indices for simplicity. $\Lambda$ denotes the active set of the sparse code $\balpha$ and $|\Lambda|$ is the cardinality of the active set. $\bD_\Lambda\in\bbR^{m\times |\Lambda|}$ is the subset of dictionary consists of the active atoms. $\bI_{|\Lambda|}\in\bbR^{|\Lambda|\times|\Lambda|}$ is identity matrix and $\mathbf{1}_{|\Lambda|}\in\bbR^{|\Lambda|}$ is an all one vector.

\begin{algorithm}[h!t!]
%\small
\caption{Dictionary and parameter update for deep sparse coding network}

\begin{algorithmic}[1]
%\caption{Task driven dictionary learning using group sparse recovery}
%\fontsize{7}{7}
% \caption{Task driven dictionary learning with joint sparsity prior}

 %\SetAlgoLined
% input: Initial $\bD$ and $\bW$, training data set $(\by, \bx)$, initial step size $\rho_0$.

\Require $\{\bD^{(h)}\}_{h=1}^H$ dictionary initialized with Gaussian random noise, initial $\{\lambda_1^{(h)}\}_{h=1}^H$.  $\{\bx_i, y_i\}$ training pairs.  $t=1$.
 \While{stopping criterion not satisfied} 
 \State Randomly choose a sample pair $\{\bx_i, y_i\}$ and let $\balpha^{(0)} = \psi(\bx_i)$.
 \For{layer $h=1$ to $H$}
\State $\balpha^{{(h)}^*} \leftarrow \arg\min_{\balpha^{(h)}>\mathbf{0}}\frac{1}{2}\norm{\bx^{(h)} - \bD^{(h)}\balpha^{(h)}}_2^2+\lambda_1^{(h)}\norm{\balpha^{(h)}}_1 + \frac{\lambda_2}{2}\norm{\balpha^{(h)}}_2^2.$ 
\NoNumber where $\bx^{(h)}=\psi(\balpha^{(h-1)})$.
 \EndFor
 
\For{ layer $h=H$ down to $1$}
\State Update the dictionary/regularization parameters with a gradient descent/projection step
\begin{align*}
\bD^{(h)} &\leftarrow \bD^{(h)} - \rho_t(\partial L/\partial \bD^{(h)}+\mu\bD^{(h)}), \\
\lambda_1^{(h)} &\leftarrow \left(\lambda_1^{(h)} - \rho_t(\partial L/\partial \lambda^{(h)}+\mu\lambda_1^{(h)})\right)_+,
\end{align*}
\NoNumber where $\rho_t$ is the learning rate at time $t$.
\EndFor 
 
\State $t\leftarrow t+1.$
\EndWhile \\
\Return $\{\bD^{(h)}, \lambda^{(h)}\}_{h=1}^H$.

\end{algorithmic}
\label{algorithm:dl}
\end{algorithm}

\textbf{Differentiation of $\partial L/\partial \bD$. } We first derive the differentiation $\partial \balpha/\partial d_{jk}$ for a single dictionary element $d_{jk}$. The inactive atoms are not updated since the  desired gradient on which $\alpha_j=0$ is not well defined \cite{MairalTDDL,Yang} and $\partial \balpha_{\Lambda^c}/\partial d_{jk}=\mathbf{0}$, where $\Lambda^c$ is the complementary of $\Lambda$. Differentiate both sides of (\ref{eq:explicit}) with respect to $d_{jk}$ for all $j\in\Lambda$:
\begin{align}
\left(\bD_\Lambda^\top\bD_\Lambda+\lambda_2\bI_{|\Lambda|}\right)\frac{\partial \balpha_\Lambda}{\partial d_{jk}}  + \frac{\partial \bD_\Lambda^\top\bD_\Lambda}{\partial d_{jk}}\balpha_\Lambda - \frac{\partial \bD_\Lambda^\top\bx}{\partial d_{jk}} = \mathbf{0},
\label{eq:before_active}
\end{align}%
which is equivalent with
\begin{align}
\frac{\partial \balpha_\Lambda}{\partial d_{jk}} &=(\bD_{\Lambda}^\top\bD_{\Lambda}+\lambda_2\bI_{|\Lambda|})^{-1}(\frac{\partial\bD^\top_\Lambda\bx}{\partial d_{jk}} - \frac{\partial \bD^\top_\Lambda\bD_\Lambda}{\partial d_{jk}}\balpha_\Lambda ).
\label{eq:grad_dictionary_L}
\end{align}%
 We reach the updating rule for a single dictionary element:
\begin{align}
\frac{\partial L}{\partial d_{jk}} &=\left(\frac{\partial L}{\partial \balpha}\right)_\Lambda^\top\cdot (\bD_\Lambda^\top\bD_\Lambda+\lambda_2\bI_{|\Lambda|})^{-1}(\frac{\partial\bD^\top_\Lambda\bx}{\partial d_{jk}} - \frac{\partial \bD^\top_\Lambda\bD_\Lambda}{\partial d_{jk}}\balpha_\Lambda ).
\label{eq:grad_dictionary_L}
\end{align}%
Stacking all elements ${\partial L}/{\partial d_{jk}}$ into ${\partial L}/{\partial \bD}$ and applying algebraic simplification:
\begin{align}
\frac{\partial L}{\partial \bD} = -\bD\bgamma\balpha^{\top} + \left(\bx-\bD\balpha\right)\bgamma^{\top},
\label{eq:dictionary_update}
\end{align}
where $\bgamma_\Lambda = (\bD_\Lambda^\top\bD_\Lambda+\lambda_2\bI_{|\Lambda|})^{-1}\cdot \partial L/\partial \balpha_{\Lambda}$ and $\bgamma_{\Lambda^c} = \mathbf{0}$. Due to the sparsity constraint, only few atoms are activated in each layer and $|\Lambda|$ is small enough for efficiently implementation (\ref{eq:dictionary_update}) on modern GPUs. 

\textbf{Differentiation of $\partial L/\partial \lambda$. }
Differentiating both sides of Eq. (\ref{eq:explicit}) with respect to $\lambda_1$:
\begin{align}
 \bD_\Lambda^\top\bD_\Lambda\frac{\partial \balpha}{\partial \lambda_1} =-1,
\label{eq:grad_lambda}
\end{align}%
and we reach at
\begin{align}
\frac{\partial L}{\partial \lambda_1} &=\left(\frac{\partial L}{\partial \balpha}\right)_\Lambda^\top\cdot -(\bD_\Lambda^\top\bD_\Lambda+\lambda_2\bI_{|\Lambda|})^{-1}=-\bgamma.
\label{eq:grad_lambda_L}
\end{align}

\textbf{Differentiation of $\partial L/\partial \bx$. }
The gradient of sparse code $\balpha$ with respect to each input signal element $\bx$ can be reached by differentiating both sides of (\ref{eq:explicit}) with respect to $x_i$:
\begin{align}
(\bD_{\Lambda}^\top\bD_{\Lambda}+\lambda_2\bI_{|\Lambda|})\frac{\partial \balpha_\Lambda}{\partial x_i} - \frac{\partial \bD_\Lambda^\top\bx}{\partial x_i} = \mathbf{0},
\label{eq:grad_input}
\end{align}
where $x_i$ is the $i^{\text{th}}$ element of $\bx$. (\ref{eq:grad_input})  is equivalent with
\begin{align}
\frac{\partial \balpha_\Lambda}{\partial x_i} &=(\bD_{\Lambda}^\top\bD_{\Lambda}+\lambda_2\bI_{|\Lambda|})^{-1}\frac{\partial \bD^\top_\Lambda\bx}{\partial x_i}. 
\label{eq:grad_input_L}
\end{align}
Therefore we have
\begin{align}
\frac{\partial L}{\partial x_i} &=\left(\frac{\partial L}{\partial \balpha}\right)_\Lambda^\top\cdot (\bD_{\Lambda}^\top\bD_{\Lambda}+\lambda_2\bI_{|\Lambda|})^{-1}\frac{\partial \bD^\top_\Lambda\bx}{\partial x_i},
\label{eq:grad_input_L}
\end{align}
which can be further simplified as
\begin{align}
\frac{\partial L}{\partial \bx} = \bD\bgamma.
\label{eq:dictionary_update}
\end{align}

\end{document}